\documentclass[runningheads]{llncs}
\usepackage[T1]{fontenc}
\usepackage{graphicx}
\usepackage{hyperref}
\usepackage{amsmath}
\usepackage{amssymb}
\usepackage{multirow}
\usepackage{booktabs}
\usepackage{xcolor}
\usepackage{orcidlink}
\usepackage{ulem}
\usepackage{authblk}
\usepackage[final]{changes}
\begin{document}
\title{Dual-LoRA and Quality-Enhanced Pseudo Replay for Multimodal Continual
Food Learning}
%
%
\author{Xinlan Wu\inst{1}\orcidlink{0009-0003-2764-4279} \and
Bin Zhu\inst{2}\orcidlink{0000-0002-9213-2611} \and
Feng Han\inst{1} \and
Pengkun Jiao\inst{1} \and
Jingjing Chen$^{\dagger}$\inst{1}\orcidlink{0009-0009-8457-8670}}
\authorrunning{X. Wu et al.}
%
\institute{College of Computer Science and Artificial Intelligence, Fudan University, China \\
\email{\{23210240332, 23210240172, pkjiao23\}@m.fudan.edu.cn}, \email{chenjingjing@fudan.edu.cn} \and
Singapore Management University, Singapore \\
\email{binzhu@smu.edu.sg}}
\maketitle              
\footnotetext[1]{$\dagger$ Jingjing Chen is the corresponding author.}
\markright{}
\begin{abstract}
Food analysis has become increasingly critical for health-related tasks such as personalized nutrition and chronic disease prevention. However, existing large multimodal models (LMMs) in food analysis suffer from catastrophic forgetting when learning new tasks, requiring costly retraining from scratch. To address this, we propose a novel continual learning framework for multimodal food learning, integrating a Dual-LoRA architecture with Quality-Enhanced Pseudo Replay. We introduce two complementary low-rank adapters for each task: a specialized LoRA that learns task-specific knowledge with orthogonal constraints to previous tasks' subspaces, and a cooperative LoRA that consolidates shared knowledge across tasks via pseudo replay. To improve the reliability of replay data, our Quality-Enhanced Pseudo Replay strategy leverages self-consistency and semantic similarity to reduce hallucinations in generated samples. Experiments on the comprehensive Uni-Food dataset show superior performance in mitigating forgetting, representing the first effective continual learning approach for complex food tasks.

\end{abstract}
\section{Introduction}
\deleted{
As global awareness of health and nutrition grows, food analysis has emerged as a critical area in AI-driven healthcare applications, from meal planning and dietary management to chronic disease prevention.
As a result, the scope of food computing has evolved from basic categorization \cite{bossard2014food} to complex tasks like ingredient recognition \cite{chen2016deep}, recipe generation \cite{chhikara2024fire} and nutrition estimation \cite{thames2021nutrition5k}.
}

\deleted{
In the literature, most existing works \cite{bossard2014food,chen2016deep,salvador2017learning} predominantly explore these tasks independently. 
}

\added{Food analysis is crucial for health-related applications like personalized nutrition. While research has evolved from basic categorization \cite{bossard2014food} to complex tasks such as ingredient recognition \cite{chen2016deep,chen2020study} and recipe generation \cite{chhikara2024fire,liu2025retrieval}, most approaches tackle these tasks in isolation.}
Multitask learning using large multimodal models has been recently studied \cite{yin2023foodlmm} for various food tasks \added{, offering a unified solution}. While the results are promising, it is trained using only static datasets and thus struggles to adapt to new food-related tasks without overwriting previously acquired knowledge \cite{zhai2024investigating}. Due to the phenomenon of catastrophic forgetting \cite{mccloskey1989catastrophic}, the model's performance on previous tasks deteriorates sharply when adapted to learn new tasks. This lack of continual learning ability renders them inefficient and impractical in real-world scenarios where data and task distributions evolve continuously. Retraining models from scratch for every new task is computationally expensive and requires long-term access to historical data—an unrealistic assumption in many settings.

Continual learning (CL) has emerged as a promising paradigm for models to acquire knowledge sequentially while mitigating catastrophic forgetting \cite{yu2024recent}. While explored for large language models (LLMs) and large multimodal models (LMMs), current approaches focus predominantly on classification tasks\cite{srinivasan2022climb,zheng2023preventing}, limiting their applicability to generative food analysis tasks like recipe generation \cite{chhikara2024fire,liu2025retrieval}. To be more comprehensive, the current multimodal continual learning (MMCL) taxonomy delineates five learning scenarios \cite{yu2024recent}, including class-incremental (CIL), domain-incremental (DIL), and task-incremental learning (TIL) adapted from traditional CL, along with two MMCL-specific paradigms, generative domain-incremental learning (GDIL) and modality-dynamic task-incremental learning (MDTIL). While these frameworks provide structured methodologies for sequential learning, fundamental limitations persist: CIL/DIL/GDIL enforce restrictive output formats (e.g., fixed label spaces or vocabulary structures), whereas TIL/MDTIL necessitate explicit task identification during inference, an artificial constraint incompatible with natural human-model interaction.

We propose a novel Dual-LoRA framework with Quality-Enhanced Pseudo Replay to address multimodal continual learning across diverse food tasks. Our approach uniquely handles varying task types - from ingredient recognition (multi-class classification) to recipe generation (text generation) - without requiring task labels. The core innovation lies in the Dual-LoRA architecture, which employs two independent low-rank adapters (LoRAs) \cite{hu2022lora}: a specialized LoRA that captures task-specific knowledge and a cooperative LoRA that preserves shared culinary knowledge across all learned tasks. To prevent knowledge interference, we introduce orthogonal regularization between the new task's specialized LoRA and previous tasks' cooperative LoRA subspaces, effectively mitigating catastrophic forgetting while maintaining model plasticity.
Complementing this architecture, we develop a pseudo replay strategy that leverages the model itself to generate historical task samples. Recognizing the hallucination challenges in LMMs \cite{bai2024hallucination}, we implement task-specific quality enhancement: for ingredient recognition, we apply majority voting with confidence thresholding across multiple generations, while for recipe generation, we select the most semantically consistent output via pairwise cosine similarity analysis. These refined pseudo samples, combined with current task data, are used to train the cooperative LoRA, creating a synergistic learning loop that simultaneously strengthens both task-specific and shared knowledge representations.

The main contributions can be summarized as follows:
\begin{itemize}
    \item To the best of our knowledge, this is the first work that achieves multi-task continual learning in food analysis. We propose a novel Dual-LoRA continual learning framework along with a new pseudo sample quality enhancement scheme to address catastrophic forgetting.
    \item We conduct experiments on Uni-Food, a food dataset containing diverse complex food-related information. The results demonstrate that our method achieves significant improvements over existing approaches in continual learning for complex food-related tasks.
    \item Through systematic ablation experiments, we empirically validate the efficacy of our architectural decisions and quantify the individual contributions of each component to the model's performance.
\end{itemize}

\section{Related Work}


\subsection{Low-Rank Adaptation}
As model sizes grow exponentially, traditional full-parameter fine-tuning becomes impractical. Parameter-Efficient Fine-Tuning (PEFT) address this challenge by selectively updating or introducing a small subset of parameters.
\deleted{Low-Rank Adaptation (LoRA) \cite{hu2022lora} has emerged as a prominent PEFT approach for large transformer-based models. By constraining weight updates to low-rank decompositions, LoRA achieves substantial parameter reduction while preserving model performance. Building upon this, O-LoRA \cite{wang2023orthogonal} introduces orthogonality constraints inspired by Orthogonal Gradient Descent (OGD) \cite{farajtabar2020orthogonal}, projecting new task gradients into subspaces orthogonal to previous tasks. Within this subspace, the neural network efficiently learns the new task while ensuring minimal alteration to the model’s output on prior tasks.}
\deleted{However, strict orthogonality may hinder beneficial knowledge sharing in food-related tasks. Our Dual-LoRA architecture incorporates both orthogonal constraints and knowledge consolidation for more flexible multi-task learning.}
\added{
Low-Rank Adaptation (LoRA) \cite{hu2022lora} is a prominent PEFT method that constrains weight updates to low-rank decompositions, achieving efficient adaptation. O-LoRA \cite{wang2023orthogonal} extends this by enforcing orthogonality between tasks' gradient subspaces to reduce interference. However, strict orthogonality may impede beneficial knowledge transfer. Our Dual-LoRA architecture incorporates both orthogonal constraints and knowledge consolidation for more flexible continual learning.
}

\subsection{Food Analysis}
\deleted{Advances in computer vision and diverse food datasets have expanded the scope of food analysis research tasks. Food-101 \cite{bossard2014food} established food category benchmarks, facilitating the development of food classification methods. VIREO Food-172 \cite{chen2016deep} leveraged the link between ingredient recognition and food classification to predict ingredients from images. Recipe generation also emerged as a key task; FIRE 
\cite{chhikara2024fire}, for example, applied BLIP to generate captions and T5 to produce recipes from both captions and ingredients. Nutrition estimation remains particularly challenging. Thames et al. introduced Nutrition5k \cite{thames2021nutrition5k}, a large-scale and deeply annotated nutritional dataset, along with a method combining images and depth information to estimate nutrients.}

\deleted{Historically, each task was studied separately with dedicated models, limiting efficiency and cross-task synergy. The rise of large language models (LLMs) and large multimodal models (LMMs) has simplified multi-task learning. Recently, FoodLMM \cite{yin2023foodlmm} proposed a versatile food assistant based on a multimodal foundation model, capable of food recognition, ingredient identification, recipe generation, nutrition estimation, and food segmentation. To our knowledge, it is the only existing LMM in the food domain. However, when learning a new food analysis task, FoodLMM still needs to train from scratch, incurring significant computational and temporal costs. In contrast, our method equips the model with continual learning capabilities, allowing our food-oriented LMM to incrementally acquire new tasks with minimal overhead.}

\added{With the release of comprehensive food datasets such as Food-101 \cite{bossard2014food}, VIREO Food-172 \cite{chen2016deep}, Recipe1M \cite{salvador2017learning}, and Nutrition5k \cite{thames2021nutrition5k}, research in food analysis has expanded from basic classification tasks to more complex challenges including ingredient recognition \cite{chen2016deep,chen2017cross,chen2020study}, recipe generation and retrieval \cite{chhikara2024fire,zhu2019r2gan,liu2025retrieval,song2024enhancing}, and nutrition estimation \cite{thames2021nutrition5k,qi2025advancing}. Early approaches typically addressed each task in isolation, developing dedicated models for specific problems. However, the emergence of large language models (LLMs) and large multimodal models (LMMs) has enabled more unified solutions. Notably, FoodLMM \cite{yin2023foodlmm} represents the first food-specialized LMM capable of handling multiple tasks—such as food recognition, ingredient identification, recipe generation, nutrition estimation, and segmentation—within a single model. Despite its versatility, FoodLMM still requires full retraining when adapting to new tasks, which is computationally expensive and impractical for continual learning scenarios. In contrast, our work equips food-oriented LMMs with continual learning capabilities, allowing incremental acquisition of new tasks with minimal overhead.}

\subsection{Continual Learning}
\label{sec: related work, continual learning}
The primary challenge in continual learning is catastrophic forgetting \cite{mccloskey1989catastrophic}, where models lose previously learned knowledge when trained on new tasks. Existing CL methods can be broadly categorized into \textit{regularization-based}, \textit{replay-based}, \textit{optimization-based}, \textit{representation-based}, and \textit{architecture-based} approaches \cite{wang2024comprehensive}. We focus on two types relevant to our work: \textit{replay-based} and \textit{optimization-based}.

According to the content of the replay, \textit{replay-based} approach can be further categorized into three sub-types: experience replay, generative replay and feature replay. Experience replay \cite{chaudhry2018efficient} stores some old training samples in a small memory buffer. Generative replay \cite{shin2017continual,ostapenko2019learning} usually requires training an additional generative model (e.g., GAN or VAE) to produce synthetic data from previous tasks. Feature replay \cite{liu2020generative,toldo2022bring} improves efficiency by replaying feature-level distributions. In our approach, we adopt a pseudo-replay strategy that leverages the LMM's own generative capability, avoiding the need for an external model.

\textit{Optimization-based} methods like OWM \cite{zeng2019continual} and OGD \cite{farajtabar2020orthogonal} constrain gradient updates to avoid interference with previous tasks. OLoRA \cite{wang2023orthogonal} utilizes low-rank adaptation subspaces and holds the different tasks' gradients orthogonal to each other. In general, the gradient projection approaches explicitly constrain the parameter update direction. We employ this type of idea when training the specialized LoRA.

\added{
While established CL methods (e.g., EWC, LwF) excel in classification, they are ill-suited for our heterogeneous task stream. Methods like EWC rely on a fixed output space to estimate parameter importance, which is absent when transitioning between multi-label recognition (ingredient recognition) and generative tasks (recipe generation). Similarly, LwF requires overlapping output spaces for effective distillation. Our Dual-LoRA framework circumvents these limitations by operating on parameter subspaces (Specialized LoRA) and replaying original task objectives (Cooperative LoRA), making it uniquely applicable to diverse multimodal tasks.
}

\section{Method}
\subsection{Preliminaries}

\begin{figure*}[t]
  \centering
  \includegraphics[width=\textwidth]{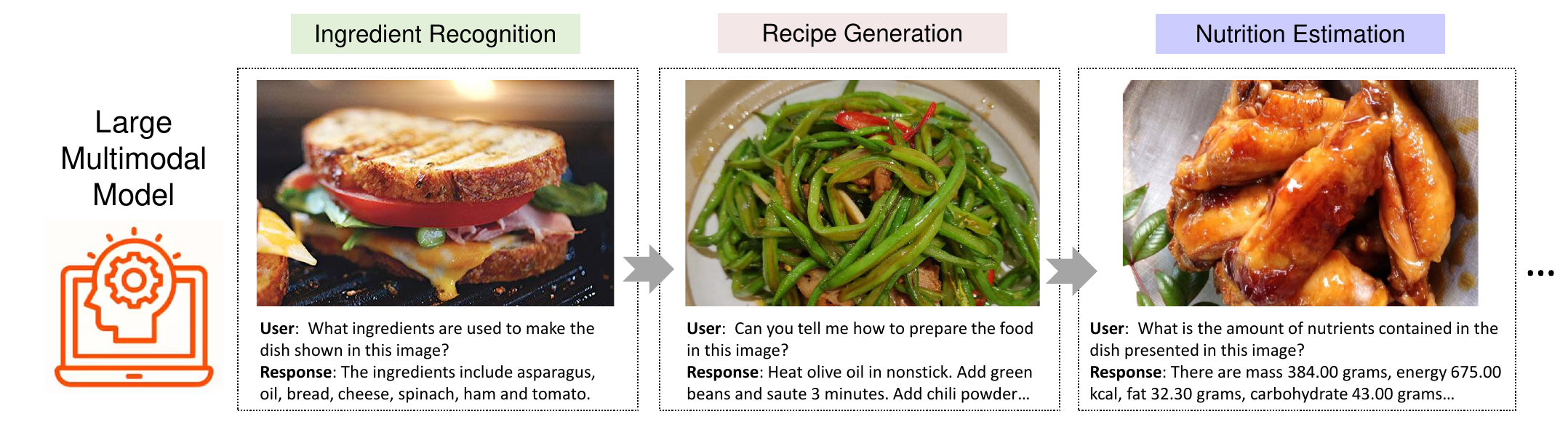}
  \caption{Our Task Learning Stream. The large multimodal model (LMM) is trained continuously across three food-related tasks: first identifying ingredients from images, then generating cooking instructions, and finally estimating nutritional content. This follows a strict task-incremental setting, meaning no historical data is reused when learning new tasks.}
  \label{fig:task_stream}
    \vspace{-6mm}
\end{figure*}

\noindent\textbf{Task Definition.} This paper investigates continual visual question answering (VQA). Contrasting with conventional offline training paradigms that assume full access to static datasets, we examine a continual learning framework where models process sequentially arriving, non-stationary data streams. To structure this analysis, we partition the dataset into $M$ distinct subtasks defined by question-type, as shown in {\autoref{fig:task_stream}}. Each subtask $T_t = \{(x_i^t, y_i^t)_{i=1}^{N_t}$\} contains its specific training dataset, where $x_i^t \in X_t$ represents the input instruction containing a food image and a question, and $y_i^t \in Y_t$ denotes the corresponding output. In the training phase, this task aims to make the LLM learn distinct VQA tasks in sequence, acquiring knowledge of the current task while memorizing historic ones. As for the inference stage, given the image and question, the model is expected to provide the corresponding answer without knowing which task the sample is from.

\noindent\textbf{LoRA.} Low-Rank Adaptation (LoRA) \cite{hu2022lora} is an efficient fine-tuning approach that constrains weight updates to low-rank subspaces. For a pre-trained weight matrix $W_{\text{init}} \in \mathbb{R}^{d \times k}$, LoRA represents updates via low-rank decomposition:
\begin{equation}
    W_{\text{init}} + \Delta W = W_{\text{init}} + BA,
\end{equation}
where $A \in \mathbb{R}^{r \times k}, \quad B \in \mathbb{R}^{d \times r}$, and rank $r \leq \min(d, k)$. $W_{\text{init}}$ remains fixed during training while $A$ and $B$ are trainable. The forward pass becomes:
\begin{equation}
    z = W_{\text{init}}x + \Delta W x = W_{\text{init}}x + BAx.
\end{equation}
This approach captures the intrinsic low-dimensionality of model updates while adding minimal parameters (only $A$ and $B$). In our framework, following \cite{hu2022lora}, only the query and value projection matrices are adapted, maintaining the pretrained model's generalization while enabling efficient task adaptation.

\subsection{Framework Overview}
\begin{figure*}[t]
  \centering
  \includegraphics[width=\textwidth]{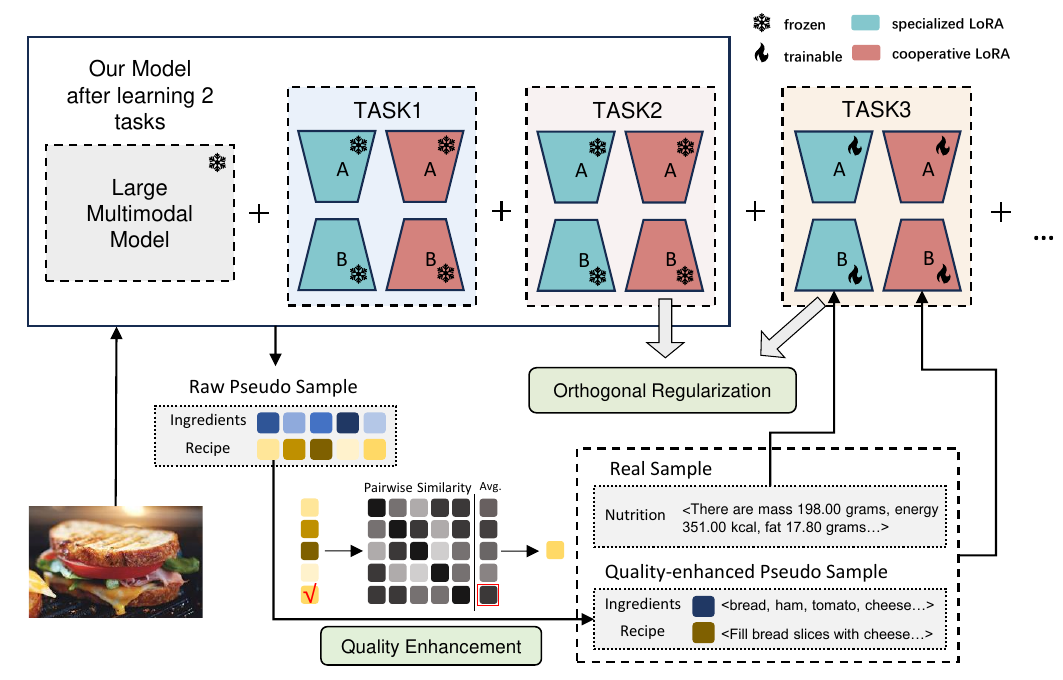}
  \caption{Our proposed architecture integrates two key components for multimodal continual food learning: (1) Dual-LoRA with specialized LoRA for task-specific knowledge (enforced via orthogonal regularization) and cooperative LoRA for shared knowledge consolidation, (2) Quality-Enhanced Pseudo Replay that leverages self-consistency and semantic similarity to generate reliable training samples from previous tasks.}
  \label{fig: framework}
    \vspace{-6mm}
\end{figure*}

The ultimate goal of continual learning is to give models the ability to continuously acquire knowledge of new tasks while retaining knowledge obtained in previous tasks. To achieve this, we propose an innovative approach that distributes two synergetic LoRA adapters per task (\autoref{fig: framework}): a \textit{specialized LoRA} for current task knowledge, and a \textit{cooperative LoRA} maintaining shared knowledge across all learned tasks. By forcing the \textit{specialized LoRA} to learn in a subspace orthogonal to the LoRA subspace associated with the previous task, we can mitigate catastrophic forgetting by preventing interference with the loss function of past tasks.

As for the pseudo-sample generation, we use the LMM in the current stage to generate the information of historical task. On this basis, some quality-enhancing processes are then performed to create the more reliable pseudo samples needed.

\subsection{Dual-LoRA}
The knowledge-sharing property among tasks in the food domain has its dual effects. On the one hand, it amplifies feature space interference during incremental learning, exacerbating catastrophic forgetting beyond conventional domain scenarios. On the other hand, cross-task knowledge transfer opportunities exist that could mutually enhance model capabilities if properly exploited. To deal with this property, we propose a Dual-LoRA architecture for each learning task: (1) A \textbf{specialized LoRA} that establishes orthogonal parameter subspaces through regularization, effectively decoupling feature representations from similar tasks. (2) A \textbf{cooperative LoRA} that leverages cross-task knowledge sharing for performance enhancement.

\noindent\textbf{Specialized LoRA.} The role of the \textit{specialized LoRA} is to specialize in learning knowledge specific to the current new task, while minimizing the negative impact on past tasks during the learning process. In the food domain, tasks typically share reusable knowledge elements. When updating gradients for a new task without considering prior tasks, the gradient space of historical tasks may still be perturbed by the new task’s updates, leading to catastrophic forgetting. Thus, we employ LoRA \cite{hu2022lora} for training and utilize its low-rank subspace as a proxy for the gradient subspace of past tasks. 

In practice, assume that the model has been trained on a continuous task stream $\{T_1, T_2, \cdots, T_{t-1}\}$, and is about to learn task $T_t$. We let our \textit{specialized LoRA} of task $T_t$ incrementally learn the new task in directions orthogonal to the \textit{cooperative LoRA} subspace of task $T_{t-1}$. This stems from the fact that the \textit{cooperative LoRA} of task $T_{t-1}$ actually encapsulates the shared subspace of all historical tasks from task $T_1$ to task $T_{t-1}$. Thus, the orthogonality between the new task and historical tasks is equivalent to the orthogonality between the \textit{specialized LoRA} of task $T_t$ and the \textit{cooperative LoRA} of task $T_{t-1}$. Following \cite{wang2023orthogonal}, the orthogonality can be expressed as:
\begin{equation}
O_t = A_{t,\text{specialized}} \cdot A_{t-1,\text{cooperative}},
\label{eq:ortho_condition}
\end{equation}
where $A$ is the low-rank decomposition matrix $A$ in LoRA. When the orthogonality equals zero, it indicates that the orthogonality is fully achieved. Thus, we define the orthogonality loss as:
\begin{equation}
L_o(A_t) = \sum_{i,j} \| O_{t}[i,j] \|^2,
\label{eq:orth_loss}
\end{equation}
where $O_t[i,j]$ denotes the element at the $i$-th row and $j$-th column of $O_t$. Therefore, our final loss function for the \textit{specialized LoRA} is defined as: 
\begin{equation}
\sum_{(x,y) \in \mathcal{D}_t} \log p_{\Theta}(y \mid x) + \lambda_o \sum_{i=1}^{t-1} L_o(A_t),
\label{eq:loss_function}
\end{equation}
where $\lambda_o$ is the weight of the orthogonality loss.

\noindent\textbf{Cooperative LoRA.} The \textit{cooperative LoRA} is designed to aggregate and preserve the shared knowledge across all tasks encountered by the model so far. A relatively robust and straightforward approach to achieve this goal is through data replay \cite{rolnick2019experience}. The replay-based methods reuse past task examples when learning new skills. There are two main types of replay-based approach: \textit{direct replay} typically retains a limited number of older training instances in episodic memory, while \textit{pseudo replay} employs generative models to learn the data distribution from previous stages and subsequently replays generated data during current task training \cite{yu2024recent}. For the reason that the previous samples are not always accessible in practice, \textit{direct replay} has to reserve some memory to save the samples, which incurs memory consumption and is not completely continual. Also, given that LMM is inherently a generative model, it can simultaneously take the responsibility of both answering questions and generating pseudo samples. Thus, we choose to use \textit{pseudo replay}.

More precisely, the model is instructed to generate pseudo samples of task $T_1$ to $T_{t-1}$ before fine-tuning on the new task $T_t$. After quality enhancement for the pseudo samples (detailed in \autoref{sec:quality enhancement}), we obtain the final pseudo samples. These samples are then mixed with real samples from task $T_t$ to train the \textit{cooperative LoRA} of task $T_t$. It is worth noting that the training objective of the \textit{cooperative LoRA} is inherited from the pretrained LMM model, with only the cross entropy loss contained, being different from \textit{specialized LoRA}'s training objective.

\subsection{Quality Enhancement for Pseudo Sample}
\label{sec:quality enhancement}
As discussed in previous section, in order to implement replay-based training method on the \textit{cooperative LoRA}, it is necessary for the model to generate reliable and high-quality pseudo samples of previous tasks. However, LMMs suffer from the issue of hallucination, and will occasionally produce outputs visually inconsistent with the input image \cite{bai2024hallucination}. As a result, directly using generated pseudo samples may lead to suboptimal results. First, raw generated samples might fail to retain critical information in historical tasks. Moreover, they might even introduce erroneous samples that can confuse the \textit{cooperative LoRA} during training. Thus, to enhance the robustness of pseudo samples, we introduce quality enhancement techniques to refine the generated pseudo samples before training.

Using self-consistency principles, we apply majority voting to select reliable replay samples. For ingredients, we first prompt the model to repeatedly generate ingredient lists for the same food image for $n$ times, resulting in a set $P=\{P_1, P_2, \cdots, P_n\}$, where each subset $P_i$ represents the ingredients generated in the $i$-th iteration. To filter out infrequent or redundant items (e.g., imperceptible seasonings), we set a threshold $t$, retaining only ingredients that appear more than $t$ times as high-confidence pseudo labels. 
The formula is as below:
\begin{equation}
    P_{enhanced} = \left\{ u \ \Big| \ \left|\{ i \mid u \in P_i \}\right| \geq t \right\},
\label{eq:ingredient improvement}
\end{equation}
where $u$ denotes every single ingredient element generated.

For pseudo recipe instruction sample quality enhancement, we also make the model to generate recipe instructions for $n$ times first, yielding a set $Q=\{Q_1, Q_2, \cdots, Q_n\}$, where each subset $P_i$ corresponds to the $i$-th generated recipe. Subsequently, we employ a Sentence-BERT \cite{reimers2019sentence} to map each recipe into an embedding vector. Then, we compute pairwise cosine similarity between all embeddings, and construct an $n \times n$ matrix, in which each row represents the consistency of the current prediction with all others (excluding diagonal elements). By averaging the consistency values per row, we derive a confidence score for each recipe. Finally, the recipe instruction with the highest confidence score is selected as the quality-enhanced recipe pseudo sample, as illustrated below:
\vspace{-0.5\baselineskip}
\begin{equation}
    Q_{enhanced} = \arg \max_i \left( \frac{1}{n-1} \sum_{\substack{j=1 \\ j \neq i}}^n \frac{Emb(Q_i) \cdot Emb(Q_j)}{\|Emb(Q_i)\| \|Emb(Q_j)\|} \right),
\label{eq:recipe improvement}
\end{equation}
where $Emb(\cdot)$ denotes the embedding function of Sentence-BERT.

\section{Experiments}

\subsection{Experimental Setting}
\begin{figure}[h]
    \centering
    \begin{minipage}{0.7\textwidth}
        \vspace{-3mm}
    \includegraphics[width=0.98\linewidth]{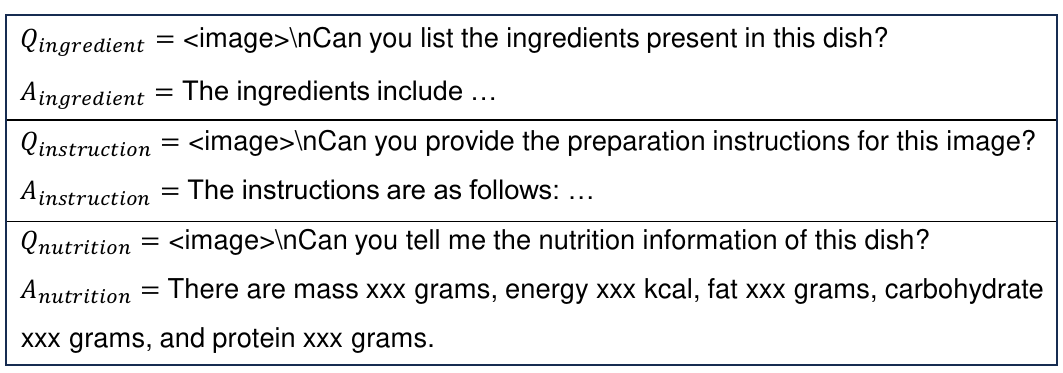}
        \vspace{-3mm}
    \caption{Samples of Data Templates.}
    \label{fig: data_templates}
    \end{minipage}
\end{figure}

\noindent\textbf{Dataset.}
We use the Uni-Food dataset \cite{jiao2024rode} for experiments. It is the only dataset with unified and comprehensive annotations for category, ingredients, recipes, and nutrition information, containing up to 99,524 food images. 
The dataset splits include 96,270 training and 3,254 testing samples. 
We organize the dataset in VQA (Visual Question Answering) \cite{antol2015vqa} format, pairing food images with QA tuples (ingredients, recipes, nutrition). Visual inputs anchor the data, while textual prompts (e.g., "List the ingredients in this dish") guide responses. Training serializes these into a unified conversational format as shown in \autoref{fig: data_templates}.

\noindent\textbf{Metrics.} Following previous works \cite{chhikara2024fire,yin2023foodlmm}, we adopt IoU (Intersection over Union) as the metric to evaluate the  quality of the generated ingredients. Meanwhile, we adopt SacreBLEU and RougeL as metrics to estimate the instruction quality of the generated recipes.

\vspace{0.4em}
\noindent\textbf{Implementation Details.}
Our Dual-LoRA structure can be implemented in any transformer-based model. 
\deleted{
In our experiments, we use LLaVA model \cite{liu2023visual} as the base model, employing the pretrained weights from LLaVA-v1.5-7B. 
We utilize standard AdamW optimizer
as a learning rate scheduler. Following O-LoRA \cite{wang2023orthogonal}, we set the orthogonality constraint weight to be 0.5 ($\lambda_o=0.5$). The number of generated raw pseudo sample $n$ for both ingredient and instruction is set to be 5, and the threshold $t$ for ingredient is 4.}
\added{
We use LLaVA-v1.5-7B as the base model. The orthogonality constraint weight $\lambda_o$ is set to 0.5. For pseudo replay, we generate n=5 samples and set the ingredient threshold t=4. The model is trained with the AdamW optimizer at an initial learning rate of 2e-4.
}
The replay proportion of our main experiment is 5\%, the results of different proportion and the reason why we choose 5\% in our main experiment can be seen in \autoref{sec: ablation}. \deleted{The learning rate is initially set to 0.0002.} All of our experiments were conducted on a machine equipped with 4 NVIDIA GeForce RTX 4090 and were implemented using DeepSpeed repository.

\subsection{Performance Comparison}

\subsubsection{Quantitative Results}
\label{sec:quan}

\begin{table*}[t]
    \centering
    \footnotesize
    \setlength{\tabcolsep}{3pt} 
    \renewcommand{\arraystretch}{1.1}
    \caption{Performance Comparison for Multimodal Continual Food Learning}
    \label{table: main}
    \begin{tabular}{@{}l c *{3}{c} *{3}{c}@{}}
        \toprule
        \multirow{2}{*}{Method} & 
        \multirow{2}{*}{\shortstack{TASK1 \\ Ing. Recog.}} & 
        \multicolumn{3}{c}{TASK2 Recipe Gen.} & 
        \multicolumn{3}{c}{TASK3 Nutrition Est.} \\
        \cmidrule(lr){3-5} \cmidrule(lr){6-8}
        & & IoU & BLEU & Rouge\_L & IoU & BLEU & Rouge\_L \\
        \midrule
        LLaVA-LoRA \cite{liu2023visual} & 36.95 & 12.73 & 6.13 & 29.29 & 0.00 & 0.00 & 0.00 \\
        \addlinespace
        LLaVA-OLoRA \cite{wang2023orthogonal} & 36.56 & 35.24 & 6.35 & 29.51 & 0.00 & 0.00 & 0.00 \\
        \addlinespace
        Ours & \textbf{38.54} & \textbf{37.29} & \textbf{6.60} & \textbf{29.83} & \textbf{36.99} & \textbf{6.36} & \textbf{28.91} \\
        \bottomrule
    \end{tabular}
        \vspace{-6mm}
\end{table*}

\autoref{table: main} presents the performance of our proposed method.
\deleted{As previously discussed, most continual learning research on LMM has focused on classification tasks. Given that our task stream exhibits unique complexity not addressed by existing continual learning methods, we selected both the base model we use and its improved variant as baselines for comparison.}
\added{
Due to the reason explained in \autoref{sec: related work, continual learning}, we mainly compare our method against the base model (LLaVA-LoRA) and its most relevant adaptation (O-LoRA) to fairly evaluate our architectural contributions.
}

Since our approach is built upon LLaVA \cite{liu2023visual}, one of the state-of-the-art LMMs, we first use LoRA to train the original LLaVA model and analyze its performance on our task stream. The results demonstrate significant catastrophic forgetting \cite{mccloskey1989catastrophic} when sequentially training LLaVA on three complex food analysis tasks. Specifically, after learning TASK2 (recipe generation), LLaVA achieved reasonable instruction metrics (SacreBLEU and Rouge\_L), indicating successful acquisition of recipe generation capabilities. However, its ingredient IoU score dropped dramatically by 65.55\%, from 36.95 to 12.73.
Subsequently, when trained on TASK3 (nutrition estimation), the model showed strong performance in nutritional prediction. Yet this came at the cost of nearly complete loss of both ingredient recognition and recipe generation abilities.
Whether questioned about ingredients or recipes, it exclusively provides nutritional information in response.

Since O-LoRA \cite{wang2023orthogonal} is an effective approach for multi-task continual learning but is only applicable to LLM, we deploy its structure on LLaVA, resulting in the LLaVA-OLoRA model. Experimental results show that after learning TASK2 (recipe generation), LLaVA-OLoRA maintains strong performance in ingredient recognition without the sharp decline observed in the original LLaVA, only declining by 2.82\%. However, after learning TASK3 (nutrition estimation), the phenomenon of catastrophic forgetting still inevitably occurs, indicating that LLaVA-OLoRA still lacks the capability to achieve continual learning on complex tasks.

Our model demonstrates markedly superior performance compared to LLaVA and LLaVA-OLoRA. In terms of ingredient recognition, the model exhibits only a 3.24\% decrease in ingredient IoU after training on TASK2 (recipe generation), followed by a mere 0.80\% reduction when subsequently trained on TASK3 (nutrition estimation). In terms of recipe generation, while LLaVA and LLaVA-OLoRA show significant performance degradation in instruction metrics after TASK3  (nutrition estimation) training, our model maintains robust performance with only 8.94\% and 3.59\% decreases in SacreBLEU and Rouge\_L respectively. These results provide compelling evidence that our model possesses unique continual learning capabilities for handling complex food analysis challenges.

\added{
In terms of efficiency, our method maintains parameter efficiency by introducing only two small LoRA adapters per task. The total parameter increase is linear and marginal (e.g., $\sim$0.2\% for three tasks). The primary overhead is pseudo-sample generation during replay, resulting in a training time approximately 1.3x that of naive LoRA fine-tuning—a reasonable trade-off for preventing catastrophic forgetting without storing raw data.
}

\subsubsection{Qualitative Results}
\label{sec:qualitative results}

\begin{figure}[h]
    \centering
        \vspace{-6mm}
    \includegraphics[width=\linewidth]{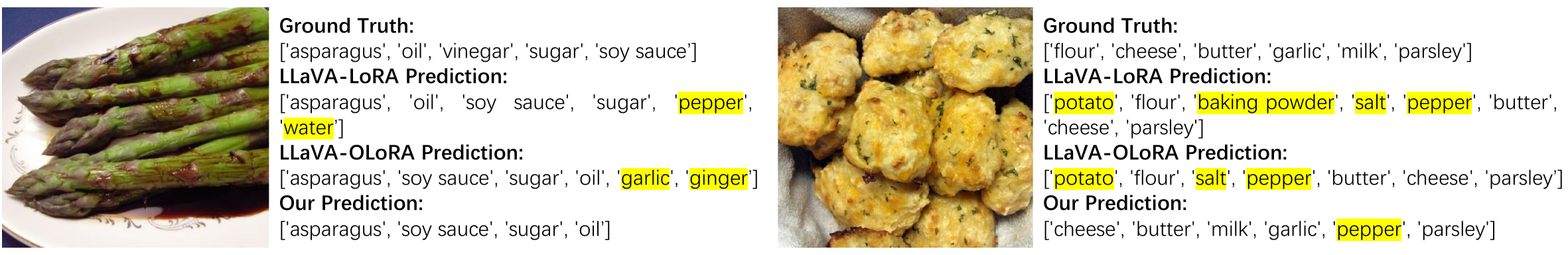}
        \vspace{-3mm}
    \caption{Comparison of ingredient recognition capability. Incorrectly generated ingredients are highlighted in yellow.}
    \label{fig:ingredient comparison}
        \vspace{-6mm}
\end{figure}

\autoref{fig:ingredient comparison} presents a qualitative comparison of ingredient recognition capability between our proposed model, baseline models (LLaVA-LoRA and LLaVA-OLoRA), and ground truth ("GT"), using every model's best-performing version immediately after TASK1 (ingredient recognition) training. The visual results demonstrate our model's superior accuracy in ingredient prediction compared to alternatives.

In the left example, while LLaVA-LoRA falsely detected "pepper" and "water", and LLaVA-OLoRA hallucinated "garlic" and "ginger", our model correctly avoided such artifacts. Notably, all models failed to identify "vinegar" - an omission we consider acceptable given the inherent difficulty of visual identification and inference, even for human observers. The right example further highlights our model's advantages: it successfully recognized critical ingredients such as "garlic" and "milk" that were missed by both baselines, while LLaVA-LoRA and LLaVA-OLoRA generated erroneous predictions (e.g., misidentifying "potato" in "cheese biscuits").

These observations provide compelling visual evidence for our model's enhanced reliability in ingredient recognition, particularly in reducing hallucination while maintaining robust detection capabilities. The comparative analysis underscores two key strengths: (1) precise adherence to visually verifiable ingredients, and (2) improved resistance to common distraction patterns exhibited by existing methods.

\subsection{Ablation Study}
\label{sec: ablation}

\begin{table*}[t]
    \centering
    \footnotesize
    \setlength{\tabcolsep}{2pt} 
    \renewcommand{\arraystretch}{1.1}
    \caption{Ablation study of Dual-LoRA architecture and quality enhancement (QE) replay strategy}
    \label{table:methods ablation}
    \begin{tabular}{@{}l c *{3}{c} *{3}{c}@{}}
        \toprule
        \multirow{2}{*}{Task Stream} & 
        \multirow{2}{*}{\shortstack{TASK1\\ Ing. Recog.}} & 
        \multicolumn{3}{c}{TASK2 Recipe Gen.} & 
        \multicolumn{3}{c}{TASK3 Nutrition Est.} \\
        \cmidrule(lr){3-5} \cmidrule(lr){6-8}
        & & IoU & BLEU & Rouge\_L & IoU & BLEU & Rouge\_L \\
        \midrule
        Ours & \textbf{38.54} & \textbf{37.29} & \textbf{6.60} & \textbf{29.83} & \textbf{36.99} & \textbf{6.36} & \underline{28.91} \\
        \hspace{1em} w/o Dual-LoRA & 36.95 & 36.22 & 6.23 & 29.33 & 36.35 & 6.34 & \textbf{29.27} \\
        \hspace{1em} w/ replay, w/o QE & 36.95 & 33.41 & 6.16 & 29.26 & 31.23 & 6.12 & 28.58 \\
        \hspace{1em} w/o replay (LLaVA-LoRA) & 36.95 & 12.73 & 6.13 & 29.29 & 0.00 & 0.00 & 0.00 \\
        \bottomrule
    \end{tabular}
        \vspace{-6mm}
\end{table*}

\noindent\textbf{Effect of Dual-LoRA and quality enhancement strategy.}
\autoref{table:methods ablation} demonstrates the ablation study of our proposed Dual-LoRA architecture and quality enhancement strategy. The results indicate that the model incorporating both of our proposed methods achieves the best performance across the majority of metrics. When Dual-LoRA is not employed while retaining the quality enhancement method for replay samples, all metrics except for Rouge\_L in the TASK3 phase exhibit certain degrees of degradation. Further removal of the quality enhancement method, leaving only the traditional replay strategy, leads to more significant declines across every metric. The approach without any replay mechanism demonstrates complete catastrophic forgetting, which is consistent with the results of \cite{rolnick2019experience,shin2017continual,chaudhry2018efficient}, illustrating the importance of replay. These findings conclusively prove that both of our proposed methods are highly effective and indispensable.

\vspace{0.4em}
\noindent\textbf{Replay data proportion.}
\begin{table}[t]
    \centering
    \begin{minipage}{0.48\textwidth}
        \caption{Ablation study of the proportion of replay data.}
        \label{tabel: replay ablation}
        \vspace{1mm}
        \resizebox{\linewidth}{!}{
            \begin{tabular}{c|ccc|ccc}
            Task Stream & \multicolumn{3}{c|}{\begin{tabular}[c]{@{}c@{}}TASK2\\ Recipe Generation\end{tabular}} & \multicolumn{3}{c}{\begin{tabular}[c]{@{}c@{}}TASK3\\ Nutrition Estimation\end{tabular}} \\ \hline
            Metric      & IoU                       & SacreBLEU                    & Rouge\_L                    & IoU                       & SacreBLEU                     & Rouge\_L                     \\ \hline
            1\%         & 36.73                     & 6.51                         & 29.74                       & 36.19                     & 6.04                          & 28.22                        \\ \hline
            5\%         & 37.29                     & 6.60                         & 29.83                       & 36.99                     & 6.36                          & 28.91                        \\ \hline
            10\%        & 37.31                     & 6.63                         & 29.90                       & 37.07                     & 6.42                          & 29.06                       
            \end{tabular}
        }
    \end{minipage}
    \hfill
    \begin{minipage}{0.48\textwidth}
        \caption{Ablation study of values of $\lambda_o$.}
        \label{tabel: lambda ablation}
        \resizebox{\linewidth}{!}{
            \begin{tabular}{c|ccc|ccc}
            Task Stream & \multicolumn{3}{c|}{\begin{tabular}[c]{@{}c@{}}TASK2\\ Recipe Generation\end{tabular}} & \multicolumn{3}{c}{\begin{tabular}[c]{@{}c@{}}TASK3\\ Nutrition Estimation\end{tabular}} \\ \hline
            Metric      & IoU                       & SacreBLEU                    & Rouge\_L                    & IoU                       & SacreBLEU                     & Rouge\_L                     \\ \hline
            $\lambda_o = 0.1$         & 37.18                     & 6.38                         & 29.35                       & 36.41                     & 6.07                          & 28.66                        \\ \hline
            $\lambda_o = 0.5$         & 37.29                     & 6.60                         & 29.83                       & 36.99                     & 6.36                          & 28.91                        \\ \hline
            $\lambda_o = 1$           & 37.22                     & 6.56                         & 29.86                       & 36.77                     & 6.20                          & 29.01                        \\ \hline
            $\lambda_o = 2$           & 37.20                     & 6.39                         & 29.66                       & 36.40                     & 6.17                          & 28.96                        \\ \hline
            $\lambda_o = 5$           & 37.06                     & 6.29                         & 29.40                       & 36.49                     & 6.00                          & 28.57                       
            \end{tabular}
        }
    \end{minipage}
    \vspace{-3mm}
\end{table}
\autoref{tabel: replay ablation} presents the impact of varying replay data proportion on experimental results. The findings demonstrate a consistent performance improvement across all metrics as the replay proportion increases, indicating a positive correlation between replay data volume and model performance. Concurrently, we observe a diminishing marginal return effect – the performance gains achieved when increasing the replay proportion from 5\% to 10\% are notably smaller than those from 1\% to 5\% (e.g., TASK2 IoU: +0.02 vs. +0.56). However, this incremental improvement requires double the computational time for pseudo sample generation. Based on this cost-performance trade-off analysis, we ultimately select 5\% replay proportion as the optimal configuration for our experiments.

\vspace{0.4em}
\noindent\textbf{Orthogonal coefficient.}
Additionally, we conducted an ablation study to investigate the impact of hyperparameter $\lambda_o$ in \autoref{eq:loss_function}. As demonstrated in \autoref{tabel: lambda ablation}, the model achieves optimal performance across most metrics when $\lambda_o=0.5$ while the results using other hyperparameters are also decent. This observation indicates that our method exhibits robustness to the hyperparameters. Furthermore, these results validate the rationale for employing orthogonal regularization within our Dual-LoRA architecture.

\section{Conclusion}
In this paper, we introduced a novel continual learning framework tailored for the food analysis domain, addressing the critical challenge of catastrophic forgetting in large multimodal models. Our Dual-LoRA architecture, comprising task-specific and multitask adapters, enables efficient knowledge retention while learning new tasks. The orthogonal subspace constraint and quality-enhanced pseudo replay mechanism further ensure robust performance across sequential tasks. Extensive experiments on the Uni-Food dataset validate our method’s superiority over existing approaches, demonstrating significant improvements in preserving prior task capabilities while adapting to new ones. 

By enabling sustainable model updates without full retraining, our framework reduces computational costs and democratizes access to evolving food analysis tools, with potential applications in personalized nutrition and public health. Our approach not only advances continual learning in food analysis but also provides a scalable paradigm for other multimodal domains requiring lifelong adaptation.

\section*{Acknowledgment}
This research is supported by the Ministry of Education, Singapore, under Academic Research Fund (AcRF) Tier 1 grant (No. MSS23C018).

%
%
%
\bibliographystyle{splncs04}
\bibliography{multimodal_food}
%




\end{document}